\documentclass[11pt,a4paper]{article}
\usepackage[hyperref]{ranlp2023}
\usepackage{times}
\usepackage{latexsym}

\usepackage{graphicx}

\usepackage{microtype}
\usepackage{amsmath} 
\usepackage{amsfonts} 
\aclfinalcopy 

\title{A Practical Survey on Zero-shot Prompt Design for In-context Learning}

\author{Yinheng Li\\
  Columbia University / New York City\\
  \texttt{li.yinheng@columbia.edu} \\}

\date{}

\begin{document}
\maketitle
\begin{abstract}
The remarkable advancements in large language models (LLMs) have brought about significant improvements in Natural Language Processing(NLP) tasks. This paper presents a comprehensive review of in-context learning techniques, focusing on different types of prompts, including discrete, continuous, few-shot, and zero-shot, and their impact on LLM performance. We explore various approaches to prompt design, such as manual design, optimization algorithms, and evaluation methods, to optimize LLM performance across diverse tasks. Our review covers key research studies in prompt engineering, discussing their methodologies and contributions to the field. We also delve into the challenges faced in evaluating prompt performance, given the absence of a single "best" prompt and the importance of considering multiple metrics. In conclusion, the paper highlights the critical role of prompt design in harnessing the full potential of LLMs and provides insights into the combination of manual design, optimization techniques, and rigorous evaluation for more effective and efficient use of LLMs in various NLP tasks.
\end{abstract}

\section{Introduction}

In recent years, transformer-based language models (such as \cite{t5}, \cite{BART}, \cite{NEURIPS2020_1457c0d6}, \cite{BERT}) have emerged as a transformative force in the field of artificial intelligence, revolutionizing Natural Language Understanding(NLU) and Generation(NLG). As model size and training data have evolved, the GPT series has exhibited extraordinary capabilities in a wide range of natural language tasks by relying on a paradigm known as in-context learning. According to \cite{NEURIPS2020_1457c0d6}, in-context learning harnesses the context provided by input data to generate appropriate responses or predictions, contrasting with traditional methods that necessitate explicit task-specific training and fine-tuning on labeled datasets. In-context learning enables large language models to capitalize on vast amounts of data and adapt to various tasks in a flexible and dynamic manner. There are several categories of in-context learning, including zero-shot, one-shot, and few-shot learning. In all types of in-context learning, the key to success lies in effective prompt design, which is occasionally referred to as an "art." This survey paper aims to categorize each type of in-context learning, discuss the core principles, examine state-of-the-art design techniques, and explore recent advancements in in-context learning, with a particular focus on zero-shot discrete in-context learning.

\section{Definition}
Although there is no formal definition for prompt design optimization, we follow the principle from \cite{NEURIPS2020_1457c0d6} and provide the definition in (\ref{equ(1)}) for prompt design in in-context learning:

\begin{equation}\label{equ(1)}
P^{\star}=\underset{P}{\arg \max } \mathbb{E}_{x_i, y_i \in \mathcal{D}}[S(f_{\theta}(P, x_i), y_i)]
\end{equation}

Here, $x_i$ represents input sentences and features, while $y_i$ denotes the target labels. $\theta$ signifies the parameters for any Large Language Models (LLMs) or Pretrained Language Models (PLMs), which remain frozen in the case of in-context learning. $f_\theta$ represents the output from LLMs given input $x_i$ and prompt $P$. $S$ is a scoring function that measures the performance of the model output in relation to the ground truth label $y_i$. The objective of in-context learning (or prompt engineering) is to identify the optimal prompt $P^*$ that maximizes the score $S$ in the test distribution.

\begin{figure}[!t]
    \centering
    \includegraphics[width=0.5\textwidth]{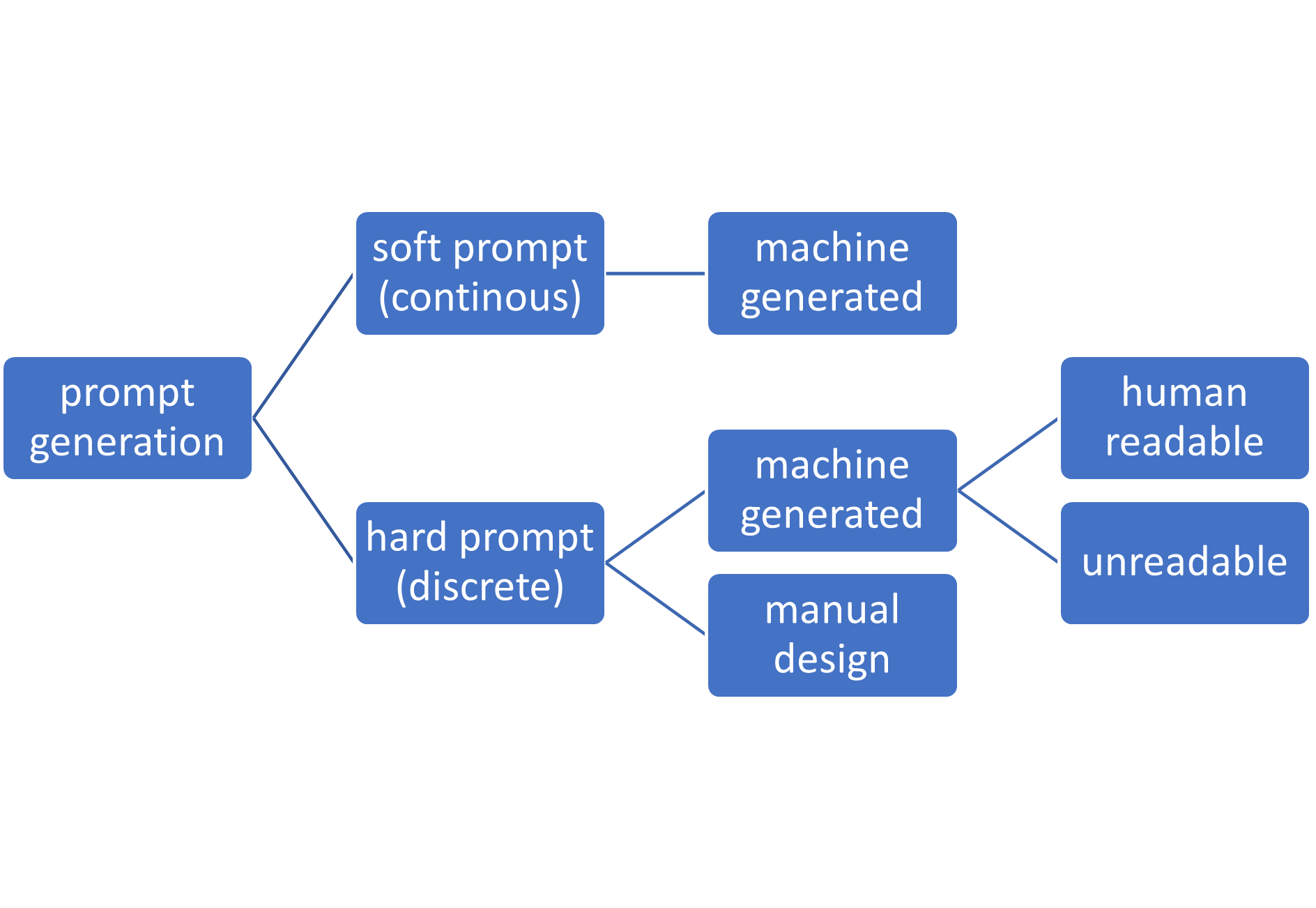}
    \caption{Prompt categorization by prompt form}
    \label{fig:picture1}
\end{figure}
Based on the structure of $P$, in-context learning can be further classified into discrete (hard) prompt when $P$ consists of a list of tokens or continuous prompt (soft) where $P$ represents an embedding vector (see Figure 1). Additionally, for zero-shot in-context learning, $P$ is independent of $x_i$, whereas for one-shot or few-shot in-context learning, $P$ can be a function of $x_i$ (from training data). This survey focuses on zero-shot in-context learning with discrete prompts and examines its application exclusively in decoder-only LLMs, such as the GPTx series.

\section{Relevant Work}
\subsection{Prompts for Encoder-only Transformer Models (BERT)}

Before the advent of in-context learning, some research efforts have been devoted to studying how to design effective prompts to enhance the performance of BERT models. As depicted in Figure 2, prompts in BERT are usually combined with input to form a cloze-style structure, while for transformer decoder-based models, prompts are more flexible.

Numerous studies have investigated prompt design in BERT. In the work by \cite{jiang-etal-2020-know}, the authors proposed heuristic-based approaches for designing discrete prompts. Dependency parsing is employed to identify useful prompts from Wikipedia. In \cite{Gao2021MakingPL}, the authors utilized T5 as a prompt generator with a beam search to create a set of diversified prompts. They then used $D_{dev}$ to select a single prompt with the best performance. In \cite{Shin2020ElicitingKF}, a gradient-based prompt search approach was proposed, wherein each prompt token is learned by directly optimizing LMs on the downstream task.

In addition to prompt designing strategies, other research work focuses on enriching the prompt candidates and ensembling the output from multiple prompts for the same input. To enrich prompts, \cite{jiang-etal-2020-know} employed back-translation to paraphrase prompts. Building on this work, \cite{Haviv2021BERTeseLT} trained a separate BERT model to rewrite prompts using the nearest BERT vector embedding.

The concept of in-context learning originates from the work by \cite{NEURIPS2020_1457c0d6}. However, BERT models can also perform similar tasks by using a single token as output. For example,
\begin{quote}
France's capital is [MASK].
\end{quote}
Only the output for the [MASK] position is used for inference. This characteristic enables the ensemble of answers from different prompts, although it is not apparent for similar practices in GPT-style models. In \cite{jiang-etal-2020-know}, the authors proposed rank-based ensemble and optimized ensemble methods to aggregate answers generated from different prompts.

Among the studies designing prompts for BERT models, the majority focus on discrete prompts (i.e., hard prompts). To the best of our knowledge, we did not find any work attempting to generate continuous prompts. In general, optimizing prompts in BERT brings only marginal improvements to the original model. Given the size and structure of BERT, it is more favorable to fine-tune on downstream tasks.
\begin{figure*}[!t]
    \centering
    \includegraphics[width=\textwidth]{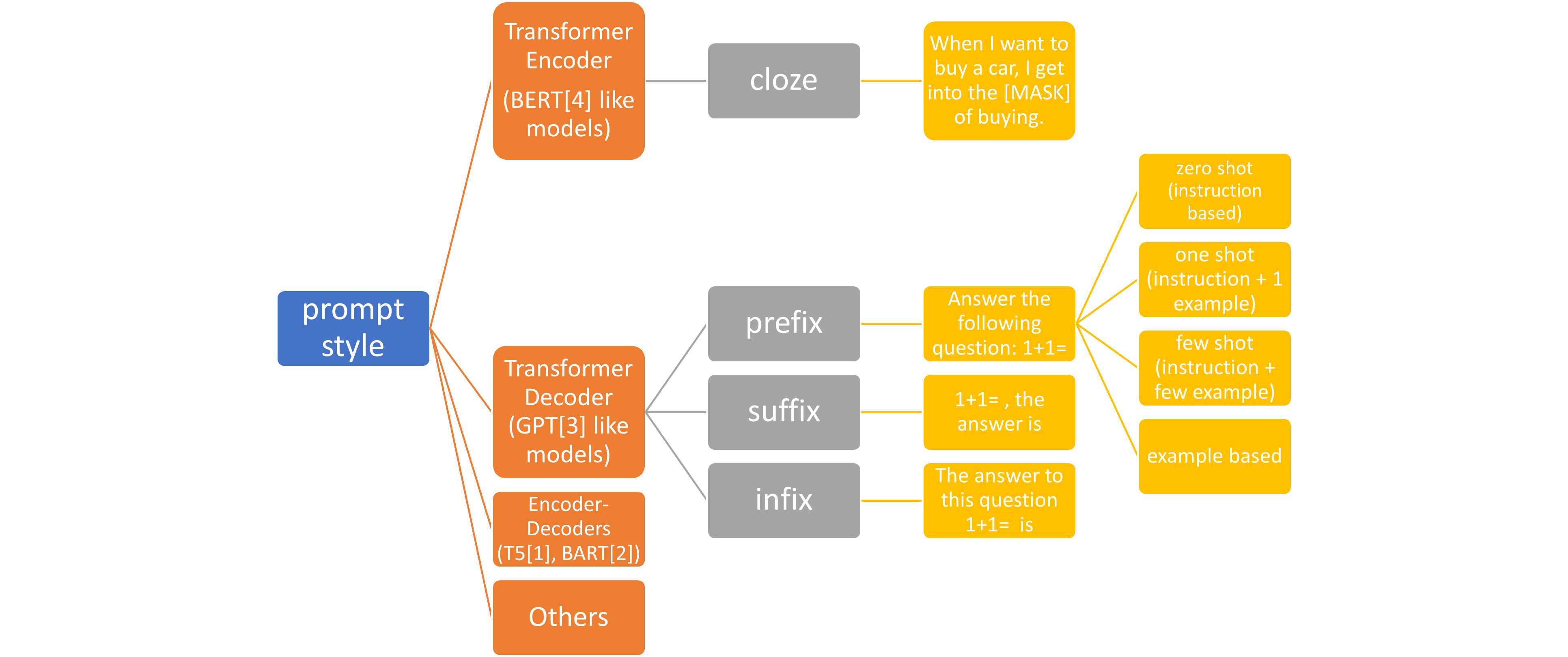}
    \caption{Prompt categorization by model types}
    \label{fig:picture1}
\end{figure*}
\subsection{Prompts for Decoder-only Transformer (GPT)}

\subsubsection{Continuous Prompt}

Another line of research has focused on optimizing soft prompts, which eliminate the constraint that prompts have to be natural language. Soft prompts can be learned and optimized directly within the same language model. The key difference between soft prompt tuning and fine-tuning is that prompt tuning typically fixes the weights of the language model and only performs gradient updates on the network that generates the prompt. Prefix-Tuning \cite{Li2021PrefixTuningOC} is one of the early works that tunes prompts on GPT-2 with a small amount of data per task, achieving comparable performance to the full data fine-tuning setting. Prefix-Tuning does not use a separate network; instead, it utilizes the same transformer network but only optimizes the input embedding of the prompt. In P-Tuning V1 \cite{GPTunderstands} and V2 \cite{liu-etal-2022-p}, the authors employ a separate LSTM network to generate the input prompt for the language model. While using soft prompts provides more flexibility in prompt design, it requires access to either the weights of language models or the ability to input vectors into language models. As recent language models are hosted as cloud services and large language models are difficult to access via vector inputs, this practice becomes less feasible when using GPT-3 or PaLM \cite{chowdhery2022palm}.

\subsubsection{Few-Shot Learning}
In the GPT paper \cite{NEURIPS2020_1457c0d6}, few-shot learning demonstrates strong NLP capabilities across various benchmarks. As the title suggests, Language Models are Few-Shot Learners. In the few-shot setting, a task description along with a few examples are presented to the model, which is then asked to complete the task for an unseen example. Numerous studies have been conducted to optimize few-shot examples and prompts to enhance performance. In \cite{Liu2021WhatMG}, the authors discovered that GPT-3 generally performs better when in-context examples are similar to the test examples. As a result, they proposed an in-context example algorithm based on example similarities. Similarity is measured using RoBERTa embedding distance in Euclidean space or cosine distance. Other works, such as \cite{Rubin2021LearningTR} and \cite{Gutierrez2022ThinkingAG}, have adopted similar example selection logic and demonstrated better performance over randomly selected examples. In addition to example selection methods, research efforts like \cite{Wu2022SelfadaptiveIL} and \cite{Kumar2021ReorderingEH} have been made to optimize the rank and order of retrieved examples.

While few-shot learning exhibits remarkable performance, according to the no free lunch(NFL) theorem \cite{Wolpert1995NoFL, 585893}, providing examples inevitably introduces bias to the prediction algorithm. In cases where out-of-distribution samples occur, applying few-shot learning can hinder the inference process.

\section{Zero-Shot Discrete Prompts}
With the recent success of Large Language Models such as GPTs, designing zero-shot discrete prompts has become increasingly popular in practice. In the experiments conducted by \cite{beyondfewshot}, the authors demonstrate that carefully engineered zero-shot prompts can actually outperform few-shot prompts. They argue that providing examples does not always help because examples tend to be interpreted as part of a narrative rather than serving as categorical guidance.

On the other hand, the advantages of using zero-shot discrete prompts can be listed as follows: (1) zero-shot prompts are highly interpretable, (2) few training data or examples are required, (3) the designing process is more straightforward as we only need to deal with task instructions, and (4) the prompt structure is flexible, allowing us to insert our input wherever needed. Zero-shot discrete prompts are also known as task instructions. There are two primary approaches to obtaining a good discrete prompt. The first is heuristic-based manual design, while the second relies on an optimization algorithm to find the optimal prompt. In this section, we focus on reviewing research on prompt design for transformer decoder style models (e.g., GPT), which has been the focus of a majority of research efforts.

\subsection{Manual Design}
In their work \cite{beyondfewshot}, the authors argue that GPT (or other LLMs) resemble a superposition of human authors. Therefore, it can be helpful to ask GPT to pretend to be a character in the prompt or use the prompt to signify a dialogue between people (i.e., task specification by memetic proxy). The authors also discuss the idea of MetaPrompts, which encapsulate a general intention that will develop towards specific meanings when additional information, such as a task question, is provided. The example prompts they provide, such as "Let's solve this problem by splitting it into steps," have been proven to be significantly helpful by subsequent works.

In the work \cite{Mishra2021ReframingIP}, the authors propose five principles for designing prompts for GPT-3 based on their observations of GPT-3's failures. These principles include: (1) using simple patterns to specify expected output, (2) using bulleted lists and assertions, (3) breaking down complex tasks into multiple simpler ones, (4) adding explicit textual statements of output constraints, and (5) customizing the instructions so that the model can directly output the results. These principles can be a good starting point for manual design.

Another line of work focuses on improving the reasoning capabilities of large language models via prompt design. The work Chain-of-Thought (CoT) \cite{cot} was initially proposed in few-shot learning, where the reasoning steps were presented as part of the solution for several few-shot examples. The zero-shot version of CoT was later proposed in \cite{Kojima2022LargeLM}, which demonstrates that inserting the single prompt "let's think step by step" into the task instruction significantly improves performance on mathematical reasoning. The authors also experimented with different templates for prompts and found that instructive prompts help improve the model's performance in mathematical reasoning, while misleading or irrelevant prompts do not contribute to performance enhancement.

\subsection{Prompt Optimization}
Finding the optimal prompt can also be treated as an optimization process, where the goal is to optimize the performance of the target task. Similar to finding the best soft prompt or finding the optimal examples for few-shot learning, algorithms can be implemented to find the best zero-shot prompt. However, such work typically requires a small set of evaluation data to assess the prompt performance. In the work by \cite{Zhou2022LargeLM}, the authors proposed Automatic Prompt Engineer (APE) for zero-shot prompt design. A LLM is used to generate a group of prompts given the task example or human description, and an iterative Monte Carlo search method is used to search for the optimal prompt given the objective function. In addition to using Monte Carlo search for prompt optimization, a gradient-free, edit-based search approach called Gradientfree Instructional Prompt Search (GRIPS) is introduced in \cite{Prasad2022GrIPSGE}. GRIPS starts from a manually designed instruction and iteratively searches among generated prompts from four operations (delete, add, swap, paraphrase) to find the optimal prompt for a target task.

Another line of research uses gradient-based methods but to generate discrete zero-shot prompts. The work FluentPrompt \cite{Shi2022TowardHR} follows the idea from AutoPrompt \cite{Shin2020ElicitingKF}, using a gradient-based method to generate discrete prompts. They also use a fluency constraint to encourage human-readable prompt outcomes, which helps improve performance. Another gradient-based prompt generation method RLPROMPT is introduced in \cite{Deng2022RLPromptOD}. This work uses a reinforcement learning structure to generate prompts that optimize the task-based reward function. The prompts generated from this framework are often incoherent gibberish but are claimed to achieve significant performance improvement.

\subsection{Evaluation}
Evaluating prompt design is very challenging. As there is no ground truth dataset for prompt generation, there is no "best" prompt but only better prompts. Therefore, the evaluation of the prompt performance for in-context learning usually falls into the following categories.

\textbf{Conditional Probability (Likelihood)}: To evaluate the performance of a text generation model, we can measure the probability of the generated text. In our case, we can calculate the conditional probability of ground truth($y$) given prompt ($p$), input($x$) or calculate the joint probability of $x, y, p$ averaging over the training data, as shown in (\ref{equ(2)})
\begin{equation}\label{equ(2)}
    \underset{x , y \in{X, Y}}{Prob(y| x,p)}
\end{equation}
This is a simple strategy because the models for in-context learning are generative language models which will generate the joint probability (likelihood) automatically. However, this metric sometimes fails to represent the actual performance of the downstream task.

\textbf{Execution Accuracy}: A more direct method to measure the performance of a prompt is to use metrics from the target task \cite{Zhou2022LargeLM}, as ultimately the performance on the task is what we care about. In addition to measuring the execution accuracy directly on the entire training set, there are ways to efficiently estimate the performance on a subset of training data to save computational cost \cite{Zhou2022LargeLM}, \cite{alphacode}.

\textbf{Prompt Transferability} is another evaluation metric reported in \cite{Zhou2022LargeLM}, \cite{Deng2022RLPromptOD} which is used to prove the quality of the prompt generation methods. However, this metric is more useful in selecting the prompt designing method than evaluating the performance of a single prompt.

\textbf{General Metrics for Language Models} should be used when using large language models via zero-shot in-context learning. It is also important to measure the performance from additional aspects. For example, if we are to build a Question-Answering system, we need to measure the risk of hallucination \cite{Ji2022SurveyOH}. If we are to build an email generation system, we may need to measure the toxicity and prevent generating any aggressive content. The work of Holistic Evaluation of Language Models (HELM) \cite{liang2022holistic} provides a great example in evaluating the performance for language models via in-context learning. 
Although various metrics have been reported in HELM for existing models, it is worth noting that the design of our prompt will directly impact the models' performance. 

\section{Conclusion}

The rapid development of large language models (LLMs) has significantly influenced various NLP tasks. Among the techniques to harness their capabilities, in-context learning with different types of prompts—discrete, continuous, few-shot, and zero-shot—has shown remarkable promise.

Discrete prompt engineering emphasizes human-readable prompts that can enhance model performance, while continuous prompt optimization involves soft prompts that can be learned and optimized directly in the same language model. Few-shot learning leverages a small number of examples to guide the model in the right direction, whereas zero-shot discrete prompts only require task instructions, offering a more straightforward design process.

Manual design of prompts can be guided by principles based on model behavior, and optimization algorithms can be used to find optimal prompts. Evaluating the performance of prompts is challenging, as there is no single "best" prompt, and various metrics need to be considered.

In conclusion, as LLMs continue to evolve, prompt design remains a crucial factor in harnessing their full potential across a wide range of applications. A combination of manual design, optimization techniques, and rigorous evaluation can lead to more effective and efficient use of LLMs in diverse NLP tasks.

\bibliographystyle{acl_natbib}
\bibliography{ranlp2023}


\end{document}